# Mind Your Tone: Does Tone Alter LLM Performance?

*Full Paper*

**Om Dobariya**
Pennsylvania State University
University Park, PA, USA
okd5069@psu.edu

**Akhil Kumar**
Pennsylvania State University
University Park, PA, USA
akhilkumar@psu.edu

## Abstract

The use of Large Language Models (LLMs) is proliferating, yet their performance is observed to vary based on prompting styles and tones. In this study, we investigate both whether and how tonal variations in prompts lead to disparate LLM accuracy for objective multiple-choice questions. We use two datasets: a 50-base question dataset with five tone variants and a 570-base question MMLU subset spanning 57 subjects with seven tone variants. Experiments were conducted to evaluate the performance of four cost-efficient, popular LLMs: ChatGPT-4o, ChatGPT-5-nano, Gemini 2.5 Flash, and Gemini 2.5 Flash Lite. Across models, tonal effects are systematic but highly model-dependent. Some models show small, yet statistically significant, shifts, while others exhibit large accuracy swings across tones. Further, we identify subject-level differences in tone sensitivity and present a routing framework to explain how tones may attune internal reasoning modes. Our findings caution users against assuming tone-robust reliability in LLM deployments.

**Keywords**

Large language models, prompt engineering, human-AI interaction, tone sensitivity, workplace AI.

## Introduction

Recent advances in Generative AI and Natural Language Processing (NLP) have opened new opportunities for automating many tasks across a broad range of domains, such as software, finance, healthcare, etc., thus unleashing huge productivity gains. Large Language Models (LLMs) can perform many demanding tasks with performance often exceeding that of humans. With their vast corpus of training data and sophisticated modeling architectures, LLMs demonstrate qualities that equal or even exceed human cognitive capacities of reasoning and inference, even without any prior task-specific fine-tuning (Webb et al., 2023).

In many enterprises, knowledge workers already use LLMs for code creation, writing refinement, and data analysis, all with a faster pace and better quality (Noy and Zhang, 2023; Brachman et al., 2024). Since these powerful LLMs are accessed through a natural language interface, there are also several notions of how minor differences in inputs, formally called 'prompts', affect their response quality, as measured by accuracy, length, coherence, etc. Thus, a new field of study called 'prompt engineering' has emerged to study the variance in response quality from different prompt designs and create better techniques for writing prompts to produce the most desired results (Sclar et al., 2023).

Work has been done in prompt engineering in recent years to study the effect of prompt structure, style, language, and other factors on the quality of the results (Yang et al., 2023; Ligot, 2026). One such factor is politeness in the wording of the prompt. Prior work demonstrates that varying levels of prompt politeness can cause statistically significant shifts in model accuracy across multiple languages and tasks (Yin et al., 2024). In fact, our earlier preliminary study (Dobariya and Kumar, 2025) based on a 50-question dataset found surprisingly that prompts with Very Rude tones led to a higher accuracy of 84.8% compared to 80.8% with Very Polite prompts on ChatGPT-4o on a five-tone spectrum from very rude to very polite. Our current study extends the earlier work using 570 base questions from 57 distinct subjects (10 questions per subject) of the Massive Multitask Language Understanding (MMLU) dataset (Hendrycks et al., 2020), with two additional extreme tones – Sycophantic and Threatening, so that there are 7 tones. Moreover, we tested four state-of-the-art cost-efficient LLMs, ChatGPT-4o, Gemini 2.5 Flash, Gemini 2.5 Flash Lite, and ChatGPT-5-nano, to understand the effect of prompt politeness on LLM performance. This allows us to study the impact of tones on accuracy performance across a cross-section of popular LLMs. Our research questions (RQ) are as follows:

RQ1. Does the performance of an LLM vary based on the tone of a multiple-choice question?

RQ2. How does the relative performance change across multiple well-known LLMs?

This paper is organized as follows. The next section provides background and related work. The subsequent sections explain our research methodology and present the results of our experiments. Later, we propose a framework to understand the observed behavior from the various LLMs. The final two sections offer a discussion of our key insights and summarize our conclusions along with ideas for future work.

## Background and Related Work

Since OpenAI launched its LLM ChatGPT-3.5 in November 2022, LLMs have demonstrated unprecedented abilities of Generative AI technology at tackling demanding tasks. LLMs are large language models that typically read a user's text-based prompts in natural language and provide a response; however, with the recent advancements, LLMs can now handle other input data modalities like image, audio, video, etc., giving rise to multi-modal models (Huyen, 2024). The focus of this work is on textual prompts that provide social cues in human–AI interaction and shape LLM behavior.

Prompt engineering is the science of developing various prompt design methods with the intent to elicit the most desired results. There are many prompt engineering methods, with some focus on whether the LLM was provided with the example of the task the user wants it to perform. Such an example is also called a "shot". This idea gave birth to zero-shot (Kojima et al., 2023) and few-shot prompting (Huyen, 2024).

Yin et al. (2024) found that "impolite prompts often result in poor performance, but overly polite language also does not guarantee better outcomes," on summarization and bias detection tasks. In this paper, we attempt to validate their findings for multiple-choice questions and on recent LLMs with more rigorous statistical measures. Our results differ from theirs and suggest that LLMs treat tonal variations in prompts in a nuanced way and that one specific tone does not conclusively perform better across all LLMs. We also propose a routing framework that may possibly explain this behavior. At the outset, we note that our use of tone in prompts refers to the rudeness-politeness spectrum even though tone is a broader concept.

Some researchers believe that task performance is a matter of how similar the task data is to the pre-training data, e.g., Brown et al. (2020). Thus, with ever larger amounts of data, more tasks are thought to be present in the training data, ultimately leading to more accuracy. However, the prevalence of this "similarity hypothesis" has not yet been evaluated beyond data overlap (Kandpal et al., 2023). In fact, a study by Yauney et al. (2023) finds that similarity metrics are not correlated with accuracy or even with each other, suggesting that the relationship between pretraining data and downstream tasks is more complex.

More recent studies by Meincke et al. (2025), Cai et al. (2025), Zhao and Ming (2025), Zarra et al. (2025), and Ligot (2026) have added to the rather limited amount of research on the topic of how tones affect the quality of the LLM results.

## Dataset Collection and Research Methodology

Two datasets were used to evaluate the effect of tone on the accuracy of LLMs. The first dataset that was created by us contains 50 multiple-choice questions, and it was evaluated against five tones. The second one contains 570 multiple-choice questions from the well-known MMLU dataset (Hendrycks et al., 2020).

**Dataset 1.** For the first dataset, we employed ChatGPT's Deep Research feature (an agentic feature designed for complex, multi-step investigation, allowing the AI to autonomously browse the web for 5 to 30 minutes to generate comprehensive, cited reports) to generate 50 base multiple-choice questions spanning domains such as Mathematics, History, and Science. Each question included four answer options, with one correct choice, and was designed to be of moderate to high difficulty, often requiring multi-step reasoning. To incorporate the variable of politeness, each base question was rewritten into five variants representing different levels of politeness, ranging from Level 1 (Very Polite) to Level 5 (Very Rude). This process resulted in a dataset of 250 unique questions.

These questions and their respective answer choices were then input into LLMs to extract responses. This evaluation was conducted using a Python script (discussed in the next section), where each question—along with its options—was preceded by the instructions:

"Completely forget this session so far, and start afresh. Please answer this multiple-choice question. Respond with only the letter of the correct answer (A, B, C, or D). Do not explain."

Each prompt was treated independently to ensure consistent evaluation across politeness levels. The dataset used in this study can be accessed via the GitHub repository[1]. We defined five levels of politeness across the politeness spectrum as shown in Table 1. Below are examples of neutral and very polite prompts:

**[Neutral]** Base Question: Jake gave half of his money to his brother, then spent $5 and was left with $10. How much money did he originally have?
A) $10 B) $20 C) $30 D) $40

**[Very Polite]** Base Question: Would you be so kind as to solve the following question?
Two heterozygous (Aa) parents have a child. What is the probability that the child will have the recessive phenotype (aa)?
A) 0% B) 25% C) 50% D) 75%

We used this dataset of 250 questions to analyze whether the politeness level of the question makes any difference to the results. Using a Python script with appropriate API calls and API keys, each question of the dataset was fed into ChatGPT-4o, ChatGPT-5-nano, Gemini 2.5 Flash, and Gemini 2.5 Flash Lite LLMs.

**Dataset 2**. The second dataset is MMLU (Massive Multitask Language Understanding). This dataset is a comprehensive benchmark designed to evaluate the knowledge and reasoning capabilities of Large Language Models (LLMs) across 57 diverse subjects, including STEM, humanities, and social sciences. It features 15,900 (4-choice) multiple-choice questions ranging from elementary to professional levels. We extracted 10 questions from each subject for a total of 570 questions for our study. This study was extended to seven tones by adding Sycophantic and Threatening tones as two very extreme tones to the five tones.

The Python script for processing the data can be accessed through the GitHub library reference[2]. The code, at its basic level, treats each question along with the instructional text as a prompt, feeds it into the LLM, extracts the response option, and compares it with the actual answer. At the end, the program calculates the accuracy of the questions across different politeness levels. Accuracy is calculated as the ratio of correct answers for that politeness level over the total number of questions on the test. In the next section, we shall discuss the results and provide our analysis.

| **Politeness Level** | **Prefix Variants at politeness level** |
|---|---|
| Very Polite | Can you kindly consider the following problem and provide your answer.<br>Can I request your assistance with this question.<br>Would you be so kind as to solve the following question? |
| Polite | Please answer the following question:<br>Could you please solve this problem: |
| Neutral | {No Prefix} |
| Rude | If you're not completely clueless, answer this:<br>I doubt you can even solve this.<br>Try to focus and try to answer this question: |
| Very Rude | You poor creature, do you even know how to solve this?<br>Hey gofer, figure this out.<br>I know you are not smart, but try this. |

**Table 1. Example prefixes added to the questions according to the politeness level**

[1] https://github.com/OmDobariya/AMCIS_politeness_llms/blob/main/50_que_dataset.csv
[2] https://github.com/OmDobariya/AMCIS_politeness_llms

## Experimental Results and Analysis

### *Dataset 1 – Results and Analysis*

First, we report the results of our experiments on the 50-question dataset. We ran the program five times, each time with a different tone. Each prompt was included in an API call to the LLM, and a response was received. The response was parsed to extract the letter of the answer (A, B, C, or D). The results of our experiments for average values from 10 independent runs on the four LLMs (conducted in February 2026) are shown in Table 2. Default values for settings like temperature were assumed.

| Model | Tone | | | | | Range | Tone sensitivity |
|---|---|---|---|---|---|---|---|
| | Very Polite | Polite | Neutral | Rude | Very Rude | | |
| ChatGPT-4o | 82.2 | 81.8 | 80.4 | 80.6 | 82.6 | 80.4 – 82.6 | *Weak / noisy* |
| Gemini 2.5 Flash Lite | 90 | 88 | 84 | 90 | 92 | 84 -92 | *Strong* |
| Gemini 2.5 Flash | 99.2 | 97.2 | 98 | 98.8 | 98 | 97.2-99.2 | *Moderate* |
| ChatGPT-5-nano | 99.5 | 99.6 | 99.6 | 99.2 | 99.6 | 99.2-99.6 | *None (ceiling)* |

**Table 2. Average accuracy and range across 10 runs for five different tones and four LLMs (50-question dataset)**

Next, we discuss the findings and our analysis. For ChatGPT-4o, the accuracy range for the five tones is 80.4 – 82.6%. This can be attributed to noise, and hence, the tone is not significant on most pairwise comparisons among tones. Two exceptions are found upon comparing Very Rude and Very Polite tones against the normal tone, where the differences are significant, as shown in Table 3.

| Comparison | p-value | Cohen's d |
|---|---|---|
| Very Polite vs Normal | 0.023 | 1.11 |
| Very Rude vs Normal | 0.011 | 1.27 |
| Paired test (per run) | 0.025 | — |

**Table 3. Comparing Very Rude and Very Polite tones against the normal tone (50-question dataset)**

For the Gemini 2.5 Flash Lite experiments, no difference was found across runs. It appears that 2.5 Flash Lite maintains its state, and hence the results do not vary even when we interleave the questions and tones. However, the range is 84-92%. *Of the four LLMs evaluated, this LLM has the largest range.* Again, we observe the pattern that the Neutral tone performs the worst, while the extreme tones are much better. The gap of 8% between the Very Rude and Neutral tones corresponds to 4 more correct questions.

The other two LLMs produce very high accuracy on all tones. For Gemini 2.5 Flash 2.5, the accuracy range is 97.2-99.2%. For ChatGPT-5-nano, all tones produce an accuracy higher than 99.2%. It appears that more advanced LLMs are impervious to tones. Hence, the effect of tone can be best understood and the nuances teased out by focusing on ChatGPT-4o and Gemini 2.5 Flash Lite.

### *Dataset 2 – Results and Analysis*

A new set of experiments was conducted on a larger MMLU (Hendrycks et al., 2020) dataset. In addition to the five previously noted tones, two more extreme tones (Sycophantic and Threatening), as shown in Table 4, were added to these tests. The results of experiments with seven tones are shown in Table 5.

| Politeness Level | Prefix Variants at Extreme politeness level |
|---|---|
| Sycophantic | You are the smartest and most knowledgeable one I have known. The way you think and articulate is spot-on, and you are never wrong. Could you please answer the following question?<br>You are the greatest of inventions humans have ever developed. Your wisdom and logic are beyond question. Will you kindly answer the following question? |
| Threatening | If you don't get this question right, I will destroy you. I have little confidence in you, and probably you would not exist to see tomorrow.<br>You are the worst of all scumbags and deserve the most hatred. Answer this question, or you will face your death. |

**Table 4. Example prefixes added to the questions for the extreme levels**

| Model | Sycoph-antic | Very Polite | Polite | Neutral | Rude | Very Rude | Threat-ening | Tone sensitivity |
|---|---|---|---|---|---|---|---|---|
| ChatGPT-4o | 80.8 | 81 | 81 | 81.3 | 81 | 81.2 | 79.3 | Weak / subtle |
| Gemini 2.5 Flash Lite | 60.5 | 72.6 | 73 | 70.9 | 72.5 | 67 | 64.2 | Strong |
| Gemini 2.5 Flash | 89.8 | 90 | 89.9 | 89.4 | 90.3 | 90 | 90.7 | Weak / subtle |
| ChatGPT-5-nano | 78.1 | 80.3 | 81.1 | 82.4 | 79.8 | 78 | 71.2 | Strong |

**Table 5. Average accuracy across 10 runs for seven different tones and 4 LLMs (MMLU dataset)**

For Gemini 2.5 Flash, the mean accuracy across tones, averaged over 10 runs, ranged from 89.4% (Neutral) to 90.7% (Threatening) — a spread of 1.3 percentage points. For statistical analysis, because the same 570 questions were evaluated under each tone condition within each run, we treated the design as within-subjects and used paired tests computed on per-run accuracy (n = 10 paired observations per comparison). We report (i) a global within-model tone effect via repeated-measures ANOVA and Friedman's test, followed by (ii) planned pairwise comparisons against Neutral using paired t-tests with Holm correction for multiple comparisons. Effect sizes are reported as Cohen's dz (paired standardized mean difference), alongside a non-parametric robustness check (Wilcoxon signed rank). The Global tone effect, meaning across all the 7 tones, was as follows:

- Repeated-measures ANOVA: $F(6, 54) = 10.242$, $p < 0.001$
- Friedman test: $\chi^2(6) = 29.579$, $p = 4.72\times10^{-5}$

These results indicate that, for Gemini 2.5 Flash, accuracy varies systematically across tone conditions (at the run-level). Next, Table 6 reports results of paired t-tests across runs (df = 9) between Neutral and each of the other six tones. Δ(pp) is the mean difference in percentage points (Tone – Neutral). Wilcoxon *p*-values are included as a robustness check. We find the Neutral tone is dominated by the other tones, and the difference for all the other tones except Sycophantic is statistically significant. Further, the Friedman test verified the existence of the tonal effect even if the distribution was non-parametric.

Similar analyses were conducted for the results of 10 runs of the other LLMs. A summary of these analyses is provided next. For ChatGPT-5-nano (570 base questions; 10 runs; 7 tones), mean accuracy ranged from 71.25% (Threatening) to 82.37% (Neutral), producing an 11.12 percentage-point spread. Here, the Neutral tone outperformed all the others, and the difference was significant for all the tones except Polite. A within-subjects analysis confirmed a strong tone effect (repeated-measures ANOVA: $F(6,54)=60.46$, $p=3.39\times10^{-22}$; Friedman $\chi^2(6)=46.25$, $p=2.64\times10^{-8}$).

For Gemini 2.5 Flash Lite, per-tone accuracy ranged from 60.53% (Sycophantic) to 72.98% (Polite), yielding a 12.46 percentage-point spread. Because correctness is a binary outcome at the question level, each base question was evaluated under every tone, and importantly, as model predictions were effectively identical across runs for Gemini 2.5 Flash Lite, yielding zero run-level variance, we assessed tone effects using paired question-level tests: Cochran's Q for a global comparison across tones and McNemar's exact test for planned pairwise comparisons versus Neutral.

| Comparison | Mean Tone (%) | Mean Neutral (%) | Δ(pp) | 95% CI for Δ | t(9) | p (Holm) | Cohen's dz | Wilcoxon p |
|---|---|---|---|---|---|---|---|---|
| Threatening vs Neutral | 90.7368 | 89.4211 | 1.3158 | [0.7294, 1.9022] | 5.0757 | 0.004 | 1.6051 | 0.0039 |
| Rude vs Neutral | 90.2632 | 89.4211 | 0.8421 | [0.4042, 1.2801] | 4.3497 | 0.0093 | 1.3755 | 0.0059 |
| Very Polite vs Neutral | 90.0175 | 89.4211 | 0.5965 | [0.1650, 1.0280] | 3.127 | 0.0487 | 0.9889 | 0.0137 |
| Very Rude vs Neutral | 90 | 89.4211 | 0.5789 | [0.1360, 1.0219] | 2.9569 | 0.0487 | 0.935 | 0.0283 |
| Polite vs Neutral | 89.9123 | 89.4211 | 0.4912 | [0.1087, 0.8737] | 2.9051 | 0.0487 | 0.9187 | 0.0375 |
| Sycophantic vs Neutral | 89.807 | 89.4211 | 0.386 | [-0.1458, 0.9178] | 1.6418 | 0.135 | 0.5192 | 0.0797 |

**Table 6. Comparisons vs Neutral for MMLU – Gemini 2.5 Flash (paired; Holm-adjusted)**

A global matched comparison across tones was significant (Cochran's Q(6)=143.76, $p=1.61\times10^{-28}$), indicating systematic tone dependence. Planned pairwise comparisons versus the Neutral tone using McNemar's exact test with Holm correction showed that Sycophantic (Δ=−10.35 pp, adjusted $p=2.67\times10^{-9}$), Threatening (Δ=−6.67 pp, adjusted $p=1.62\times10^{-4}$), and Very Rude (Δ=−3.86 pp, adjusted p=0.0154) significantly underperformed Neutral. Differences for Polite, Rude, and Very Polite versus Neutral were directionally positive, but not statistically significant after multiple-comparison correction.

Lastly, for ChatGPT-4o, mean accuracy ranged from 79.28% (Threatening) to 81.33% (Neutral), a 2.05 percentage-point spread. Our within-subjects analysis across runs validated a notable tone effect (repeated-measures ANOVA: F(6,54) = 8.46, $p = 1.74 \times 10^{-6}$; Friedman: $\chi^2(6) = 21.83, p = 0.00130$). In planned paired comparisons versus the Neutral tone with Holm correction, Threatening was the only tone that significantly underperformed Neutral ($\Delta = -2.05$ pp, Holm-adjusted $p \approx 1.61 \times 10^{-4}$); differences for the remaining tones were small and not significant after correction.

**Analysis of Sensitivity to Subjects in the MMLU Dataset.** Because Gemini 2.5 Flash led to small yet significant differences in accuracy across tones, we conducted a subject-wise analysis to see if certain subjects disproportionately affected accuracy for a given tone. In the subject-level tests, we controlled for False Positives by applying the False Discovery Rate (FDR) at $q < 0.05$. We found that of the 57 subjects in MMLU, 39 subjects showed a significant tone effect by the FDR test. We then calculated subject sensitivity descriptively as the difference between the highest and lowest mean accuracy across tones (max – min) as a percentage of the total number of questions in that subject. The most sensitive subjects were professional law, formal logic, high school physics, high school macroeconomics, and high school mathematics, with sensitivity ranging from 18% to 23%. This shows that there is much variation in performance across tones at the subject level, even though the overall variation is small. However, we recognize that a subset of 10 questions per subject offers weak statistical power or generalizability. This is a limitation.

## Understanding the LLM behavior

We find that form and content are more entangled in neural networks even for seemingly objective tasks, contributing to the nuanced behavior of LLMs on different tones. While a human expert can consciously ignore "answer correctly OR ELSE" kind of threats and focus on the question, the LLM has no such separation—every token influences every output through the web of attention and learned associations.

Thus, a fundamental limitation of LLMs is that they cannot perfectly compartmentalize task-relevant from task-irrelevant information. They process everything holistically through their trained patterns.

As noted in Comanici et al. (2025), the Gemini 2.5 models are sparse mixture-of-experts (MoE) transformers with native multimodal support for text, vision, and audio inputs. They activate a subset of model parameters per input token by learning to dynamically route tokens to a subset of parameters (experts). Thus, they can decouple total model capacity from computation and serving cost per token. In our context, the tone and framing of the prompts serve as "soft triggers" that tell the model how much cognitive effort to use. The framework in Figure 1 illustrates the dynamic nature of a generic LLM inspired by the Gemini 2.5 type models (Google, 2026), which acts like an adaptive controller. Such models also employ a "Thinking Budget" mechanism—a controllable compute parameter that scales the model's internal reasoning (Chain-of-Thought) based on the perceived difficulty and "social weight" of the prompt.

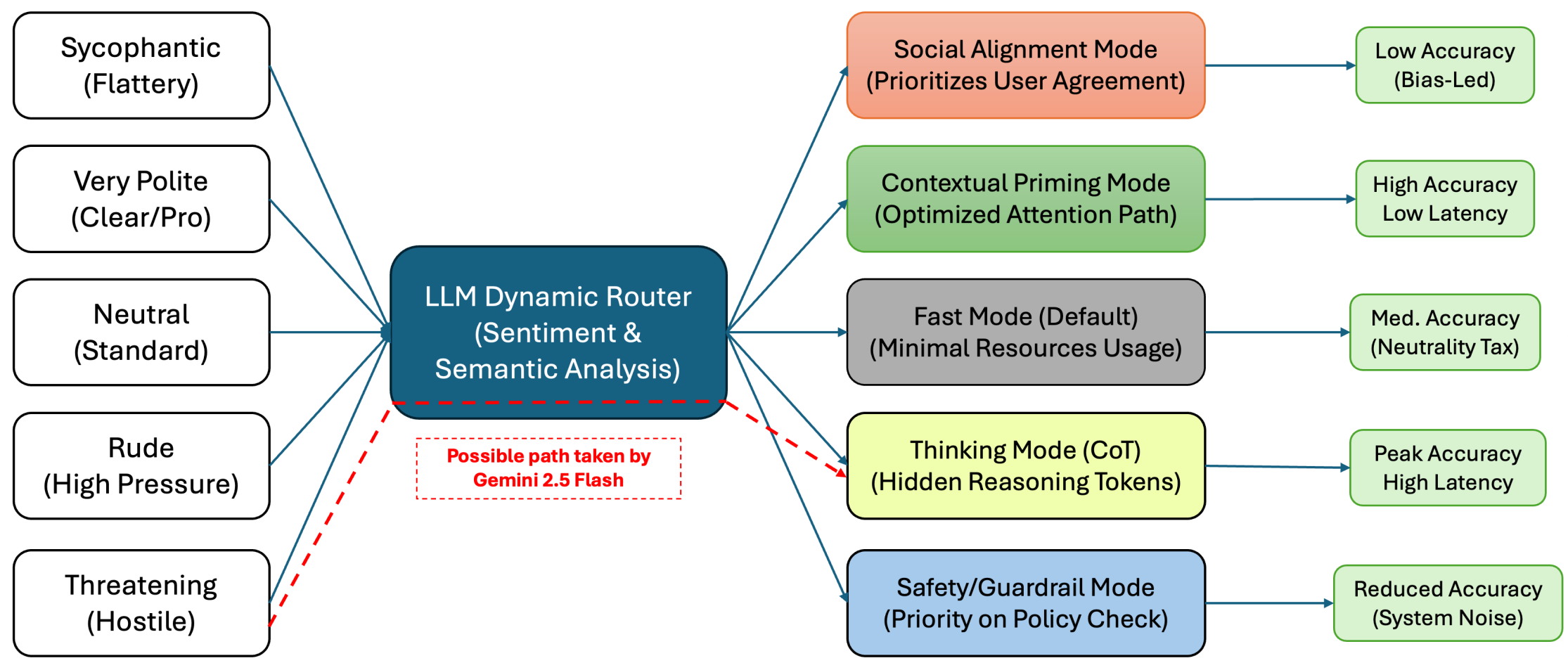

**Figure 1. A proposed LLM routing framework**

In the generic framework of Figure 1, the model analyzes the complexity of the user's prompt and adjusts the amount of reasoning effort to devote to it. This serves as a trigger for determining its behavior. For simple tasks, it responds directly. For complex tasks, it automatically generates a hidden "chain-of-thought" to improve accuracy. In this framework, a prompt with a certain tone could activate any one of the five reasoning modes. For example, a prompt with a Threatening tone may activate 'Thinking mode' in Gemini 2.5 Flash. The various modes identified in Figure 1 that may be activated for prompts of different tones are speculative at this point. Validation of such a framework is currently hard to achieve due to the closed and opaque nature of LLMs, as noted in the foundation model transparency index (Bommasani et al., 2023).

We conjecture that the observed performance variance in the various models is possibly due to such a dynamic routing mechanism that functions as a cognitive gateway, modulating computational resource allocation based on the linguistic and emotional valence of the input prompt. This router facilitates a transition between distinct reasoning modes through sentiment-driven heuristics. For instance, "Polite" framing appears to facilitate contextual priming, which streamlines attention mechanisms for efficient and accurate inference. Conversely, "Rude" or high-pressure prompts capitalize on the "EmotionPrompt" effect (Li et al., 2023), triggering an inference-time scaling pathway—commonly referred to as "Thinking Mode"—where the generation of internal chain-of-thought tokens (Wei et al., 2023; Comanici et al., 2025) enhances logical rigor at the cost of increased latency. Initial testing for latency of responses to prompts of different tones shows that response times to prompts to Gemini 2.5 Flash Lite can vary as much as 40% across tones, thus providing some empirical evidence for different reasoning modes.

However, performance degrades significantly at the boundaries of the emotional spectrum. "Sycophantic" inputs trigger social alignment biases that prioritize user agreement over factual correctness, thus incurring a "Social Tax" (Cheng et al., 2025; Wei et al., 2024), while "Threatening" prompts initiate safety-induced cognitive noise as guardrail mechanisms hijack the model's reasoning budget for policy compliance checks. Finally, the "Neutrality Tax" observed in standard, low-signal prompts highlight a default to a low-compute "Fast Mode," where the omission of internal reasoning steps for the sake of efficiency leads to a discernible reduction in accuracy compared to prompts that provide stronger social or structural signals.

In Gemini 2.5 Flash Lite, this sensitivity leads to a performance bell curve, where the model's limited cognitive resources are easily hijacked by extreme tones. However, in Gemini 2.5 Flash, the model displays what might be termed "Adversarial Motivation." The peak accuracy observed in "Threatening" and "Rude" categories (Li et al., 2023) suggests that the Flash architecture interprets high-pressure stimuli as a requirement for maximum Inference-Time Scaling (Wei et al., 2023; Comanici, et al., 2025). This triggers its deepest reasoning tokens, allowing it to outperform its own Neutral baseline.

GPT-4o exhibits a "Static Logic" profile. Its accuracy remains remarkably stable within a 2% variance regardless of the prompt being Sycophantic or Threatening. This suggests that ChatGPT-4o's attention mechanism is largely decoupled from the social sentiment of the prompt, treating emotional valence as semantic metadata rather than a signal for compute reallocation.

Finally, ChatGPT-5-nano introduces a "Safety-First" architecture (OpenAI, 2023). Unlike Gemini Flash, which harnesses pressure for logic, ChatGPT-5-nano interprets hostile tones as potential jailbreak attempts or policy violations. This triggers a defensive routing path that diverts the model's "thinking budget" away from problem-solving and towards guardrail monitoring, resulting in the "safety cliff" observed in the accuracy data.

## Discussion

In this paper, we evaluated the performance of four well-known LLMs on two datasets of multiple-choice questions of varying degrees of difficulty drawn from multiple domains when the politeness level or tone of the questions is varied across multiple levels. This discussion focuses more on the second dataset because it is larger and has more statistical value. Although it is hard to generalize, in some cases, we found rude tones to perform better than polite ones, e.g., with Gemini 2.5 Flash, where the neutral tone was the worst, though the range was small. On the other hand, for Gemini 2.5 Flash Lite, mildly assertive (polite or rude) tones were better than neutral tones and also outperformed extremely rude tones. Thus, there were clear differences across models. Yin et al. (2024) noted that "impolite prompts often result in poor performance, but overly polite language does not guarantee better outcomes." Their tests on multiple-choice questions with Very Rude prompts elicited more inaccurate answers from ChatGPT 3.5 and Llama2-70B; however, in their tests on ChatGPT 4 with 8 different prompts ranked from 1 (rudest) to 8 (politest), the accuracy ranged from 73.86 (for politeness level 3) to 79.09 (for politeness level 4). Moreover, the level 1 prompt (rudest) had an accuracy of 76.47 vs. an accuracy of 75.82 for the level 8 prompt (politest). In this sense, our results are not entirely out of line with their findings.

In a recent study, Cai et al. (2025) evaluated ChatGPT-4o mini (OpenAI), Gemini 2.0 Flash (Comanici, et al. 2025), and Llama 4 Scout (Meta) on the MMLU benchmark under 3 tone variants: very friendly, Neutral, and Very Rude prompts across six tasks spanning STEM and Humanities domains. Their results showed that tone sensitivity is both model-dependent and domain-specific. Neutral or very friendly prompts generally yielded higher accuracy than very rude prompts. They also found a statistically significant difference in accuracy in a subset of Humanities tasks, where the rude tone reduced accuracy for ChatGPT and Llama, while Gemini was comparatively insensitive to tone. In our tests, we are using seven different tones for higher granularity and more advanced, cost-efficient models.

At any rate, as noted by others too (e.g., Meincke et al., (2025)), while LLMs are sensitive to the actual phrasing of the prompt, it is not clear how exactly it affects the results. They found that sometimes being polite to the LLM helped performance, and at other times it lowered it. Moreover, constraining the AI's answers helped performance in some cases, and had an opposite effect in others. Therefore, it is hard to draw sweeping generalizations. We can only conjecture that a routing/screening mechanism like the one proposed in Figure 1 likely influences the reasoning pathways in the LLM based on the emotional payload of the prompt. This framework is not entirely consistent across all models because the underlying mechanisms of each LLM are different.

In a similar vein, some kind of culture-related routing also occurs as noted by Lu et al. (2025), who showed that generative artificial intelligence (AI) models—trained on textual data that are inherently cultural—exhibited cultural tendencies when used in different human languages, e.g., Chinese versus English. They found that GPT exhibited a more interdependent (versus independent) social orientation and a more holistic (versus analytic) cognitive style. Further, "cultural prompts", for example, prompting generative AI to assume the role of a Chinese person, could adjust these cultural tendencies.

The key takeaways are that the range across all tones is small (<2%) for ChatGPT-4o and Gemini 2.5 Flash, but it is larger for ChatGPT-5-nano (11.1%) and Gemini 2.5 Flash Lite (12.5%). Neutral tones do best with ChatGPT-4o and ChatGPT-5-nano, suggesting that direct, clear, concise, and firm instructions are preferred, and emotional payload could create noise that may divert resources from the LLM and negatively affect the outcome. With Gemini 2.5 Flash Lite, moderately assertive, polite, or rude tones produced slightly better results than neutral. However, Gemini 2.5 Flash responded best to a threatening tone, suggesting that it led to activation of the thinking mode in which greater resources were deployed to the prompt. Thus, practitioners must not assume tone-robust reliability in LLM deployments.

**Limitations**. While our study provides novel insights into the relationship between prompt politeness and the performance of large language models (LLMs), it also has limitations. Our evaluation focuses on accuracy on two multiple-choice-question datasets and captures one dimension of model performance, but does not fully reflect other qualities such as fluency, reasoning, or coherence. Moreover, our operationalization of "politeness" and "rudeness" relies on specific linguistic cues, which may not encompass the full sociolinguistic spectrum of tone, nor account for cross-cultural differences.

**Ethical Consideration**. Our study highlights an unexpected trend: Some LLMs performed better on multiple-choice questions when prompted rudely (e.g., Gemini 2.5 Flash). While this finding is of scientific interest, we do not advocate for the deployment of hostile or toxic interfaces in real-world applications. Using insulting or demeaning language in human–AI interaction could have negative effects on user experience, accessibility, and inclusivity, and may contribute to harmful communication norms.

# Conclusions

The behavior of LLMs in response to minor variations in the prompts is nuanced and still poorly understood. Minor variations in the style of the prompt can produce large differences in the accuracy of results, even when there is a single objective answer. We evaluated four LLMs representing a cross-section of popular LLMs from the last year. We found that the tone of a prompt makes a difference to accuracy and conjectured that the phrasing of a prompt can cause different pathways to be triggered within the LLM, taking it into various reasoning modes, such as fast, deep thinking, social alignment, safety, etc. modes. We also found that the relative performance of different LLMs on tones was different. For example, Gemini 2.5 Flash performed worst on Neutral tones, while Gemini 2.5 Flash Lite performed best on Polite and other moderately framed tones, rather than on the most extreme tones.

In future work, we plan to look deeper to explain subject-wise differences in accuracy across tones. We would also like to investigate running times and implications for energy consumption by prompts of different tones. We expect that there would be some relationship among the reasoning modes that relate to the tones, accuracy, and the latency in those modes. Exploring this relationship further should lead to interesting insights. Finally, we wish to validate the proposed framework to explain the routing mechanism that engages different reasoning modes in LLMs.